\documentclass{article}
\usepackage{spconf,amsmath,graphicx}
\usepackage{url}            
\usepackage{booktabs}       
\usepackage{amsfonts}       
\usepackage{nicefrac}       
\usepackage{microtype}      
\usepackage{subcaption}

\def\hyph{-\penalty0\hskip0pt\relax}
\title{Understanding Anatomy Classification\\ Through Attentive Response Maps}
\name{Devinder Kumar*$^{1}$, Vlado Menkovski*$^{2}$, Graham W. Taylor$^{3}$ and Alexander Wong$^{1}$\thanks{*Part of this work was done while affiliated to Philips Research, Eindhoven}}
\address{{}$^{1}$University of Waterloo, Waterloo, ON, Canada \hyph N2L 3G1\\
{}$^{2}$Technische Universiteit Eindhoven, Eindhoven, Netherlands\\
{}$^{3}$University of Guelph, CIFAR and Vector Institute,  ON, Canada}
\begin{document}
%
\maketitle
\begin{abstract}
One of the main challenges for broad adoption of deep learning based models such as convolutional neural networks (CNN), is the lack of understanding of their decisions. In many applications, a simpler, less capable model that can be easily understood is favorable to a black-box model that has superior performance. In this paper, we present an approach for designing CNNs based on visualization of the internal activations of the model. We visualize the model's response through attentive response maps obtained using a fractional stride convolution technique and compare the results with known imaging landmarks from the medical literature. We show that sufficiently deep and capable models can be successfully trained to use the same medical landmarks a human expert would use. Our approach allows for communicating the model decision process well, but also offers insight towards detecting biases.
\end{abstract}
\section{Introduction}
	\label{sec:intro}

Understanding the decision process of a deep neural network for classification can be challenging due to the very large number of parameters and model's tendency to represent the information internally in a distributed manner. Distributed representations have significant advantages for the capability of the model to generalize well~\cite{srivastava2014dropout}, however the trade-off is the difficulty in communicating the model's reasoning. In other words, it is difficult to represent what information was used by the model and it arrived at a particular output. In certain applications such as healthcare, understanding the decision process of a model can be a vital requirement.

\begin{figure}[t]
    \centering

        \includegraphics[trim = 0cm 0cm 0cm 0cm,height=5cm,width=1\linewidth]{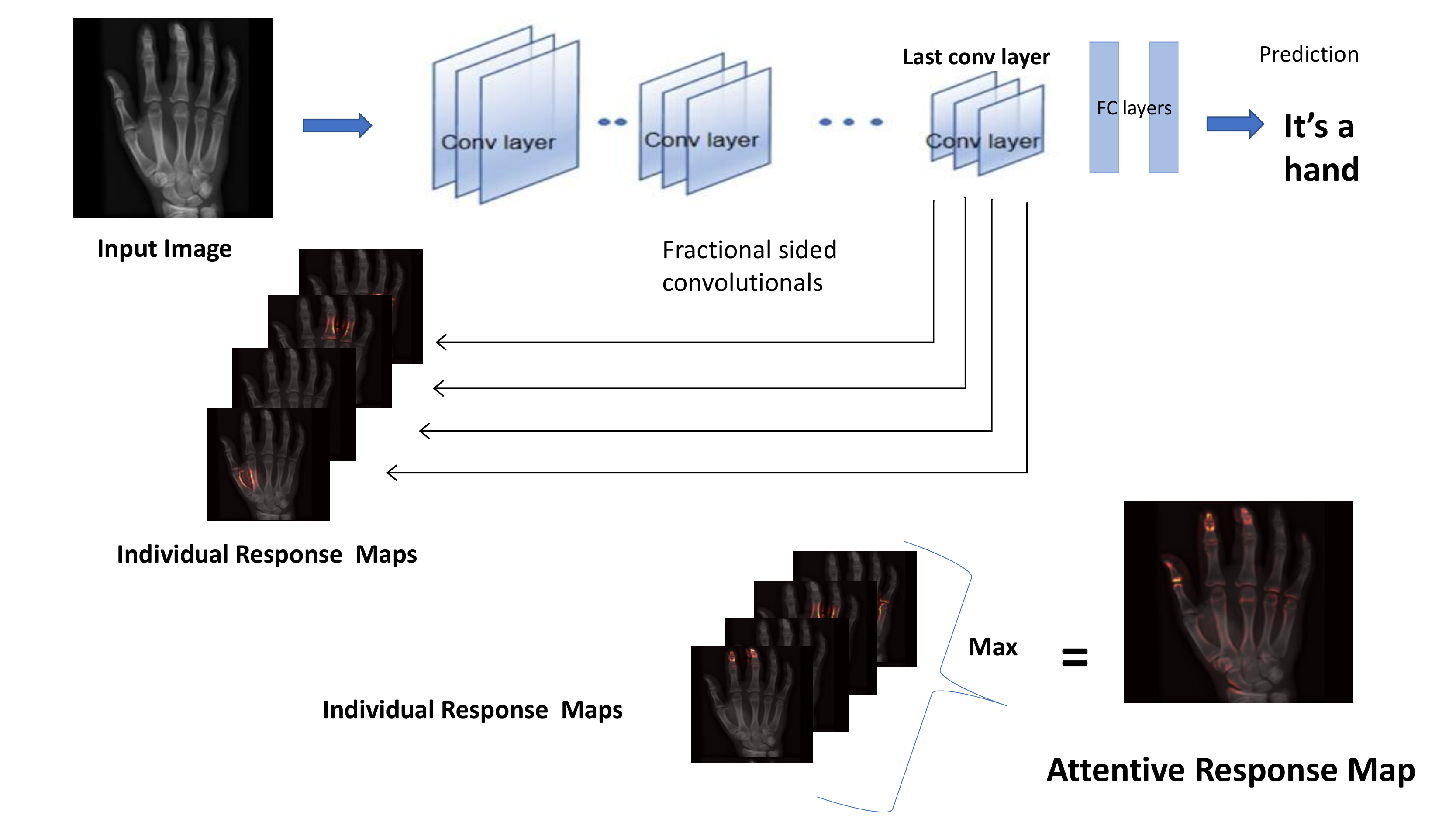}
    \caption{overview of the proposed visualization framework for understanding and visualizing human anatomy prediction. As an illustrative case in this paper, a new x-ray anatomy image is fed into the convolutional neural network to obtain the prediction (in this case, body-part classification) and through the fractionally strided convolution, individual attentive response maps are computed through top-$n$ units from the last convolutional layer in the network. By computing the max operation per pixel across all the individual maps, we obtain the attentive response map. The attentive response map shows: 1) The locations in the input image that are contributing to decision making and 2) the level of dominance of such locations.}
  \label{overview}
\end{figure}

One direction towards understanding how Convolutional Neural Networks (CNN) process the information internally is through visualization. The work of Zeiler et. al~\cite{zeiler2014visualizing}, Mahendran et. al~\cite{mahendran2015understanding}, Zhou et. al.~\cite{zhou2015cnnlocalization} etc., have shown that the inner working of the CNN can be projected back to the image space in way that is comprehensible to a human expert. We build on the work of \cite{zeiler2014visualizing} and present an approach to understand the decision making process of these networks through visualizing the information used as a part of this process as attentive response maps. Our approach is based on a fractionally strided convolutional technique~\cite{zeiler2014visualizing}, which we apply to the anatomy classification problem using X-ray images~\cite{menkovski2015can}. However, rather than examining the model over the whole dataset and trying to understand the sensitivity of the model to the data, we examine the model's response to individual data points. We also find that existing methods that present different saliency maps of the sensitivity of the model's output still do not provide a clear representation that can be used to communicate with the experts. We based our approach on visualizing attentive response maps formed through the maximally activated feature maps from the last convolutional layer in an overlayed image. This depiction provides the most informative and effective way to communicate the information from the image used in the decision process of the model. Furthermore, we compare this information to medically relevant landmarks in the images such as anatomical features that an expert would use to identify an organ. In this comparison, we find that shallow models that do not have sufficient capacity fail to use relevant landmarks. Additionally, we find that even deep models that generally perform well on test data do not necessarily use accurate landmarks. Finally, we show that adjusting training and augmentation hyperparameters based on the insight from visualizing attentive response maps leads to models that use medically relevant landmarks while attaining superior performance on test data and give indication of robustness in terms of generalization.

\section{Approach}
\label{method}
\begin{figure}[t]
\centering
\includegraphics[trim = 0cm 0cm 8cm 0cm,height=6.1cm,width=1.0\linewidth]{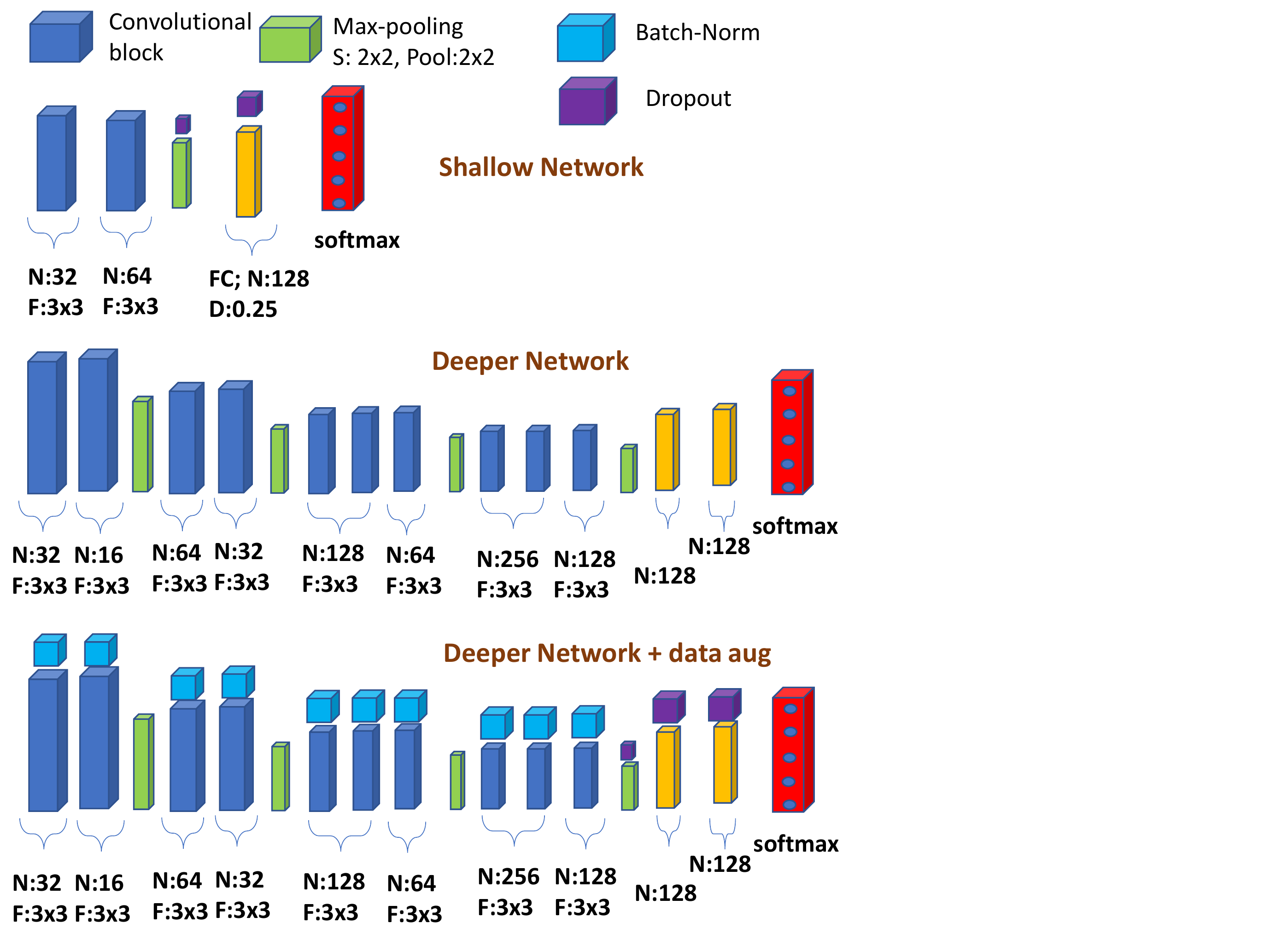}

\caption{Architecture of three different CNNs used in this study.}
  \label{fig:model_arch}
\end{figure}

In order to understand the decision making process of deep CNNs and to construct an informed approach to designing models, we build three different deep CNN models with different architectures and hyper-parameters: a shallow CNN, a deeper CNN without data augmentation and a deeper CNN with data augmentation inspired by the work of Razavian et al.~\cite{razavian2016visual}. The network architecture for each model is depicted in Fig.~\ref{fig:model_arch}. After successfully training the above mentioned networks, we examine which part of a particular input image from an anatomy class, particularly the spatially distributed information, is used in the decision process of the CNN. It is done by visualizing attentive response maps from the top $n$ most~\textit{activated} units of the last convolutional layer in the above described models, similarly to Bau et. al.~\cite{netdissect2017}. The top $n$ units are used to visualize the parts of the input image that the network considers~\textit{important}. The formation of attentive response maps are done by projecting the top unit activations back to image space. The back projection to input space is achieved by using the fractionally strided convolution, also known as the transposed convolution, and sometimes incorrectly termed the deconvolution technique~\cite{zeiler2014visualizing} as shown in Fig.~\ref{overview}. To explain the formulation for the formation of attentive response maps, let us consider a multi-layered neural network with $n$ of layers.

Recently Kumar et. al.~\cite{kumar2017explaining} proposed the CLEAR method which uses class-based maps. Even though CLEAR is quite effective in explaining CNN predictions, it can't be applied for scenarios with a large number of classes ($>10-15$). Since our method bears similarity to the CLEAR approach, we use the same notation below as explained in~\cite{kumar2017explaining} for better understanding and to highlight the important differences.

To explain the formulation of the attentive response maps, first consider a single layer of a CNN. Let $\hat h_l$ be the deconvolved output response of the single layer $l$ with $n$ unit weights $w$. The deconvolution output response at layer $l$ then can be then obtained by convolving each of the feature maps $z_{l}$ with unit weights $w_{l}$ and summing them as:

$\hat h_{l} = \sum_{k=1}^n z_{k,l} * w_{k,l} $. Here $*$ represents the convolution operation. For notational brevity, we can combine the convolution and summation operation for layer $l$ into a single convolution matrix $G_{l}$. Hence the above equation can be denoted as: $\hat h_{l} = G_{l}z_{l}$.

For multi-layered CNNs, we can extend the above formulation by adding an additional un-pooling operation $U$ as described in~\cite{zeiler2014visualizing}. Thus, we can calculate the deconvolved output response from feature space to input space for any layer $l$ in a multi-layer network as:

\begin{equation}
 R_{l} = G_{1}U_{1}G_{2}U_{2} \cdots G_{l-1}U_{l-1}G_{l}z_{l} .\
\end{equation}

For attentive response maps, we specifically calculate the output responses from individual units of the last conv. layer of a network. Hence, given a network with last layer $L$ containing $n$ top activated units, we can calculate the attentive response map; $R(\underline{x}\lvert f)$ (where $\underline{x}$ denotes the response back-projected to the input layer, and thus an array the same size as the input) for any unit $f$ (${1 \leq f \leq n}$) in the last conv layer as:

\begin{equation}
 {R(\underline{x}\lvert f)} = G_{1}U_{1}G_{2}U_{2} \cdots G_{L-1}U_{L-1}G_{L}^f z_{L} .\
\end{equation}

Here $G_{L}^f$ represents the convolution matrix operation in which the unit weights $w_{L}$ are all zero except that at the $f$\textsuperscript{th} location.

Given the set of individual attentive response maps, we then compute the dominant attentive response map, $\hat{D}(\underline{x})$, by finding the value at each pixel that maximizes the attentive response level, $R(\underline{x}\lvert f)$, across all top $n$ units:

\begin{equation}
 \hat{D}(\underline{x}) = \operatornamewithlimits{argmax}\limits_{f} {R(\underline{x} \lvert f)} .\
\end{equation}

\begin{figure}[h]
\centering
\includegraphics[trim = 0cm 0cm 0cm 0cm,height=4cm,width=0.8\linewidth]{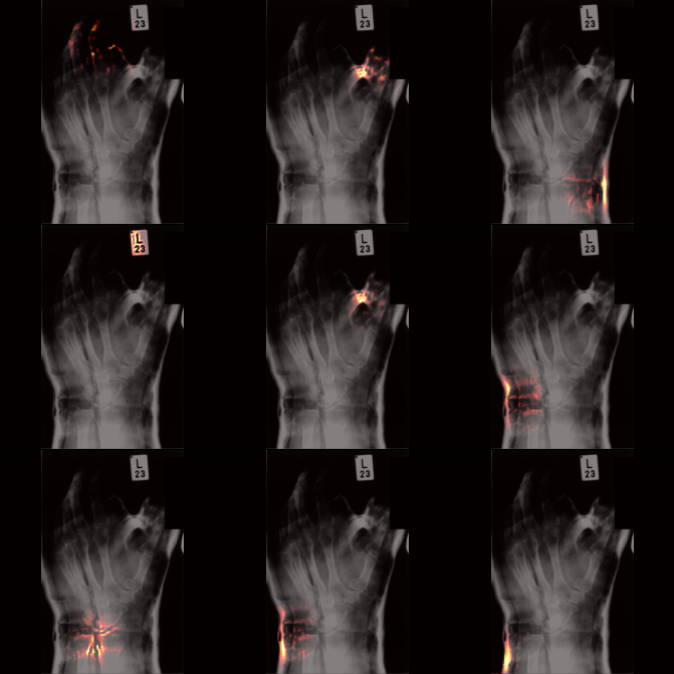}

\caption{Individual attentive response maps for top 9 activated units from the last conv layer of the deeper network with augmentation for the hand class example, mis-classified as cranium. From the figure, it is evident that the top 9 most activated units are focusing on the wrong information present in the signal. Best viewed in color.}
  \label{deep_cra}
\end{figure}

Next, we then examine the correlation of those regions obtained through visualizing the dominant attentive response maps with identified regions and shapes of image landmarks that are mentioned in the medical radiology literature. With the qualitative assessment, we can establish that the same landmarks that are described in the medical image literature are also used by the CNN. For example, we observe that the particular outlines of bones are used to detect the organ in the image rather than some background information. We use this to guide the decisions for the model architecture and learning algorithm. We can furthermore use this method to detect biases in the models. In certain examples of mis-classification (Fig.~\ref{deep_cra}), we can observe that the information used for making decisions is part of an artifact rather than the object in the image. This understanding can inform us about the possible adjustments to the pre-processing of data augmentation procedures needed to remove the bias from the model.

\begin{figure}[t]
\centering
\includegraphics[trim = 0cm 0cm 0cm 0cm,height=7cm,width=1.0\linewidth]{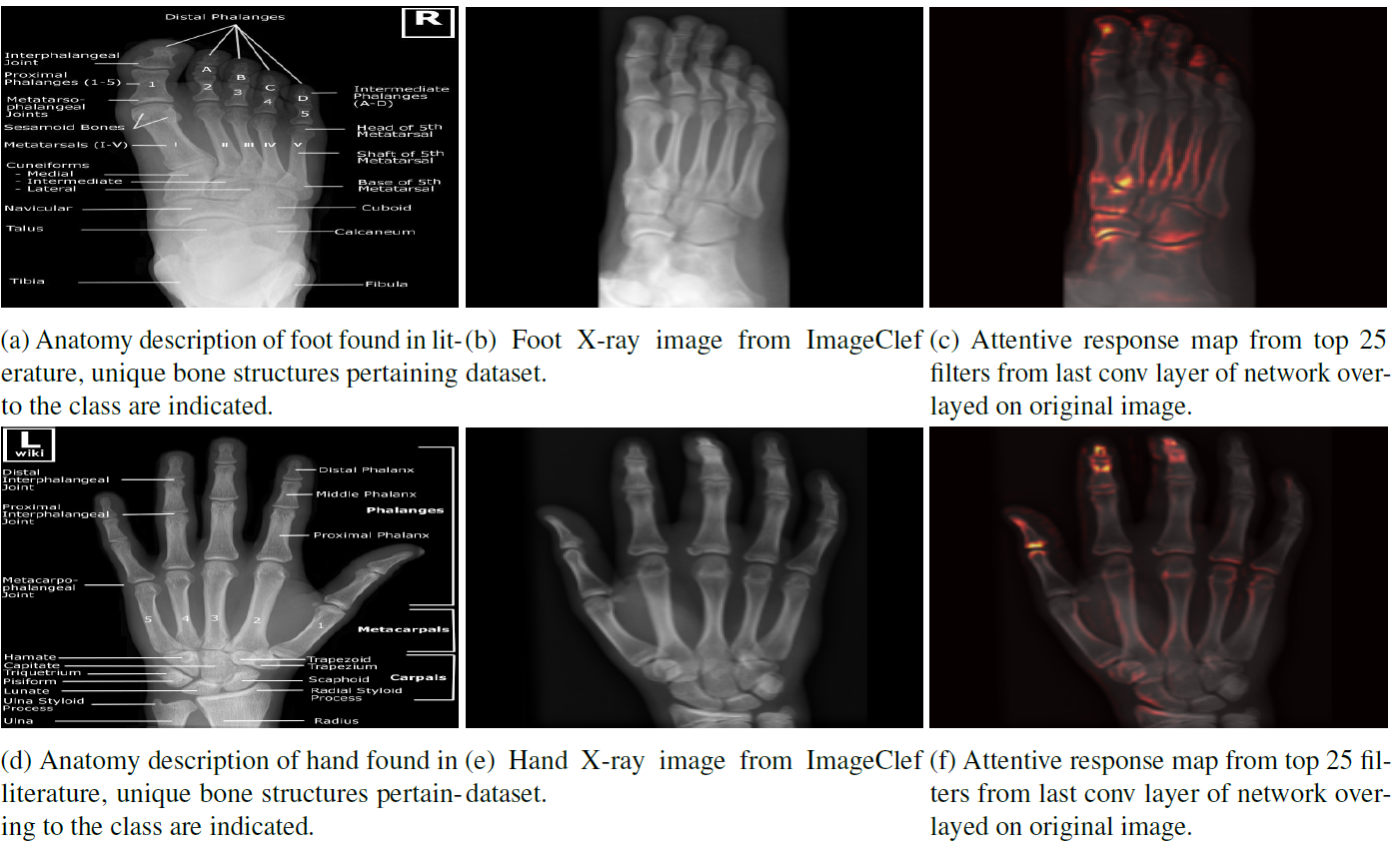}

\caption{Correspondence between anatomical descriptions found in the literature that are used by human experts ((a) \& (d)) and the attentive response maps overlayed on the original images ((b) \& (e)) from the last conv layer of the deeper network with data augmentation ((c) \& (f)) for the foot and hand class. It can be observed that the deeper neural network uses the same landmarks as a human expert for anatomy classification. Best viewed in color.}
  \label{fig:deep_hand_and_foot}
\end{figure}

\section{Experiments and Results}
To visualize and understand the decision making of a deep neural network, we used anatomy classification from X-ray images as an example use-case. To train our three different convolutional neural networks, radiographs from the ImageClef 2009 \hyph Medical Image Annotation task \footnote {http://www.imageclef.org/2009/medanno} were used. This data set consists of a wide range of X-rays images for clinical routine, along with a detailed anatomical classification. For uniform training without any bias, we removed the hierarchical class representation and removed the classes consisting of less than 50 examples. Using this, we ended up with 24 unique classes e.g.~foot, knee, hand, cranium, thoracic spine etc., from the full body anatomy.

For training the three networks described in Section~\ref{method}, we resized the images to 224 $\times$ 224. For evaluation, we divided the ImageClef dataset (14,676) images into randomly selected training and test sets with 90 \% and 10 \%  of the data respectively. For the third (deeper) network specifically, we used various data augmentation techniques ranging from cropping, rotation, translation, shearing, stretching and flipping. We trained the three networks for all the 24 classes simultaneously. The results obtained by training the three models are shown in Table~\ref{accuracy}.

\begin{table}[h]
	\centering
	\caption{Results: Accuracy in percent for three different networks trained on the ImageClef 2009 annotation task}
	\begin{tabular}{p{1.9cm} p{1.75cm}p{3.2cm}}
	\hline
	\centering
		\textbf{Shallow Net} & \textbf{Deeper Net} & \textbf{Deeper Net+data aug}\\

		\hline

			 $71.1$ 	& 	 	$90.36$				  &  	\textbf{95.62} \\
		\hline
	\end{tabular}
	\label{accuracy}
\end{table}

\begin{figure}[ht!]
    \centering
    	\begin{subfigure}[]{0.5\textwidth}
       \centering
        \includegraphics[trim = 4cm 0cm 0cm 0cm,height=3cm,width=0.45\linewidth]{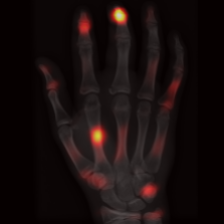}
        \includegraphics[trim = 0cm 0cm 0cm 0cm,height=3cm,width=0.45\linewidth]{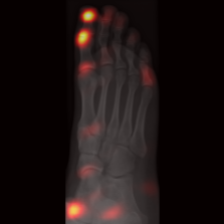}
    \end{subfigure}%
    ~

    \caption{Attentive response maps overlayed on the original images from the last conv layer of the deeper network with no data augmentation for foot and hand class. It can be observed that this network fails to use the same landmarks as a human expert for anatomy classification, as shown in Fig.~\ref{fig:deep_hand_and_foot}: (a) \& (c). Best viewed in color.}
  \label{deep_no_aug}
\end{figure}

\begin{figure*}
\centering
 \includegraphics[trim = 0cm 0cm 0cm 0cm,height=2.5cm,width=1\linewidth]{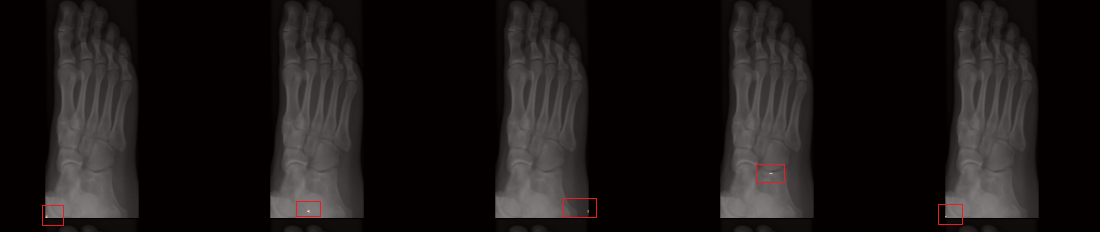}
\caption{Focus area of the top 5 attentive response maps from top $5$ most activated units from last conv layer of the shallow network. For clarity, instead of top 25 only top 5 units are shown separately. It is evident that the network doesn't learn any medically relevant landmarks. Best viewed in color.}
  \label{shallow_foot}
\end{figure*}

We visualized the internal activations of the models on test data through attentive response maps. More specifically, we combined the attentive response maps of the top $n$ = $25$ units from the last convolutional layer and overlayed them on the original image. In this way we constructed the focused attentive response maps that can be easily examined by a human expert.  The $n$ = $25$ was chosen empirically as it produced attentive response maps closer to the anatomical landmarks with least number of units. The results are shown in~\ref{fig:deep_hand_and_foot},~\ref{deep_no_aug} and~\ref{shallow_foot}  for foot and hand classes from ImageClef dataset.

In Fig.~\ref{fig:deep_hand_and_foot} we show a correspondence between the obtained attentive response maps and the anatomical landmarks from the medical literature 
\footnote{\url{http://www.meddean.luc.edu/lumen/meded/radio/curriculum/bones/Strcture_Bone_teach_f.htm}}. Particularly for the foot image, we can observe that the edges of the metatarsals' shaft has been used together with the distal phalangies, navicular, cuboid, tibia, and fibula. Similarly for the hand, three of the distal phalanxes, many of the heads of joints, metacarpals' shafts as well certain carpals. In contrast to this, in Fig.~\ref{deep_no_aug} and Fig.~\ref{shallow_foot} we can observe that the shallow and deep network trained without specific data augmentation fails to learn such specific landmarks. These models use broader ranges that are clearly not as specific as the information used in the first model. From the above visual results~\footnote{We obtained similar results for the other classes as well, but due to space constraints, only the results of two classes are shown.} as well as the performance of the final model we come to the conclusion that sufficiently deep neural network models can be successfully trained to use the same medical landmarks as a human expert while attaining superior performance.

\vspace{-0.5cm}
\section{Conclusion}
\label{future}
We propose an approach that allows for evaluating the decision making process of CNNs. We show that the design of the model architectures for deep CNN and the training procedure does not necessarily need to be a trial-and-error process, solely focused on optimizing the test set accuracy. Through attentive response map visualization, we managed to incorporate domain knowledge and overall managed to achieve a much more informed decision process, which finally resulted in a model with superior performance. This approach is applicable to many different image analysis applications of deep learning that are unable to easily leverage the potentially large amount of available domain knowledge. Furthermore, visually understanding the information involved in the model decision allows for more confidence in its performance on unseen data.

\bibliographystyle{IEEEbib}
\bibliography{references}

\begin{thebibliography}{1}

\bibitem{srivastava2014dropout}
N.~Srivastava, G.~E Hinton, A.~Krizhevsky, I.~Sutskever, and R.~Salakhutdinov,
\newblock ``Dropout: a simple way to prevent neural networks from
  overfitting.,''
\newblock {\em Journal of Machine Learning Research}, vol. 15, no. 1, pp.
  1929--1958, 2014.

\bibitem{zeiler2014visualizing}
D~Zeiler, M and R.~Fergus,
\newblock ``Visualizing and understanding convolutional networks,''
\newblock in {\em European Conference on Computer Vision}. Springer, 2014, pp.
  818--833.

\bibitem{mahendran2015understanding}
A.~Mahendran and A.~Vedaldi,
\newblock ``Understanding deep image representations by inverting them,''
\newblock in {\em IEEE CVPR}, 2015, pp. 5188--5196.

\bibitem{zhou2015cnnlocalization}
B.~Zhou, A.~Khosla, Lapedriza. A., A.~Oliva, and A.~Torralba,
\newblock ``{Learning Deep Features for Discriminative Localization.},''
\newblock {\em CVPR}, 2016.

\bibitem{menkovski2015can}
V.~Menkovski, Z.~Aleksovski, A.~Saalbach, and H.~Nickisch,
\newblock ``Can pretrained neural networks detect anatomy?,''
\newblock {\em arXiv preprint arXiv:1512.05986}, 2015.

\bibitem{razavian2016visual}
A.~S Razavian, J.~Sullivan, S.~Carlsson, and A.~Maki,
\newblock ``Visual instance retrieval with deep convolutional networks,''
\newblock {\em ITE Transactions on Media Technology and Applications}, vol. 4,
  no. 3, pp. 251--258, 2016.

\bibitem{netdissect2017}
D.~Bau, B.~Zhou, A.~Khosla, A.~Oliva, and A.~Torralba,
\newblock ``Network dissection: Quantifying interpretability of deep visual
  representations,''
\newblock in {\em CVPR}, 2017.

\bibitem{kumar2017explaining}
D.~Kumar, A.~Wong, and G.~W Taylor,
\newblock ``Explaining the unexplained: A class-enhanced attentive response
  (clear) approach to understanding deep neural networks,''
\newblock in {\em IEEE CVPR-Workshop}, 2017.

\end{thebibliography}
\end{document}